\pdfoutput=1

\documentclass[11pt]{article}
\usepackage[dvipsnames]{xcolor} 
\usepackage[]{ACL2023}

\usepackage{times}
\usepackage{latexsym}

\usepackage[T1]{fontenc}

\usepackage[utf8]{inputenc}

\usepackage{microtype}

\usepackage{inconsolata}

\usepackage{booktabs}
\usepackage{multirow}
\usepackage{adjustbox}
\usepackage{amsfonts}
\usepackage{amsmath}

\title{Gloss-Free End-to-End Sign Language Translation}

\author{{Kezhou Lin$^1$
\quad Xiaohan Wang$^1$
\quad Linchao Zhu$^1$
\quad Ke Sun$^2$
\quad Bang Zhang$^2$
\quad Yi Yang$^1$} \\
$^1$ReLER, CCAI, Zhejiang University \quad $^2$DAMO Academy, Alibaba Group\\
\texttt{kezhoulin@zju.edu.cn}
\quad \texttt{wxh1996111@gmail.com}
\quad \texttt{zhulinchao7@gmail.com} \\
\quad \texttt{xisheng.sk@alibaba-inc.com}
\quad \texttt{bangzhang@gmail.com}
\quad \texttt{yangyics@zju.edu.cn}
}
\begin{document}
\maketitle

\begin{abstract}
In this paper, we tackle the problem of sign language translation (SLT) without gloss annotations. Although intermediate representation like gloss has been proven effective, gloss annotations are hard to acquire, especially in large quantities. This limits the domain coverage of translation datasets, thus handicapping real-world applications. To mitigate this problem, we design the \textbf{Glo}ss-\textbf{F}ree \textbf{E}nd-to-end sign language translation framework (GloFE). Our method improves the performance of SLT in the gloss-free setting by exploiting the shared underlying semantics of signs and the corresponding spoken translation. Common concepts are extracted from the text and used as a weak form of intermediate representation. The global embedding of these concepts is used as a query for cross-attention to find the corresponding information within the learned visual features. In a contrastive manner, we encourage the similarity of query results between samples containing such concepts and decrease those that do not. We obtained state-of-the-art results on large-scale datasets, including OpenASL and How2Sign.\footnote{Our code and model will be available at \url{https://github.com/HenryLittle/GloFE}.}
\end{abstract}

\section{Introduction}

Sign language is a type of visual language mainly used by the community of deaf and hard of hearing. It uses a combination of hand gestures, facial expressions, and body movements to convey the message of the signer. Sign languages are not simple transcripts of the corresponding spoken languages. They possess unique grammar structures and have their own linguistic properties. According to the World Federation of the Deaf, there are over 70 million deaf people around the world. The study of automated sign language processing can facilitate their day-to-day life.

\begin{figure}[htp]
    \vspace{-0.05in}
    \centering
    \includegraphics[width=\linewidth]{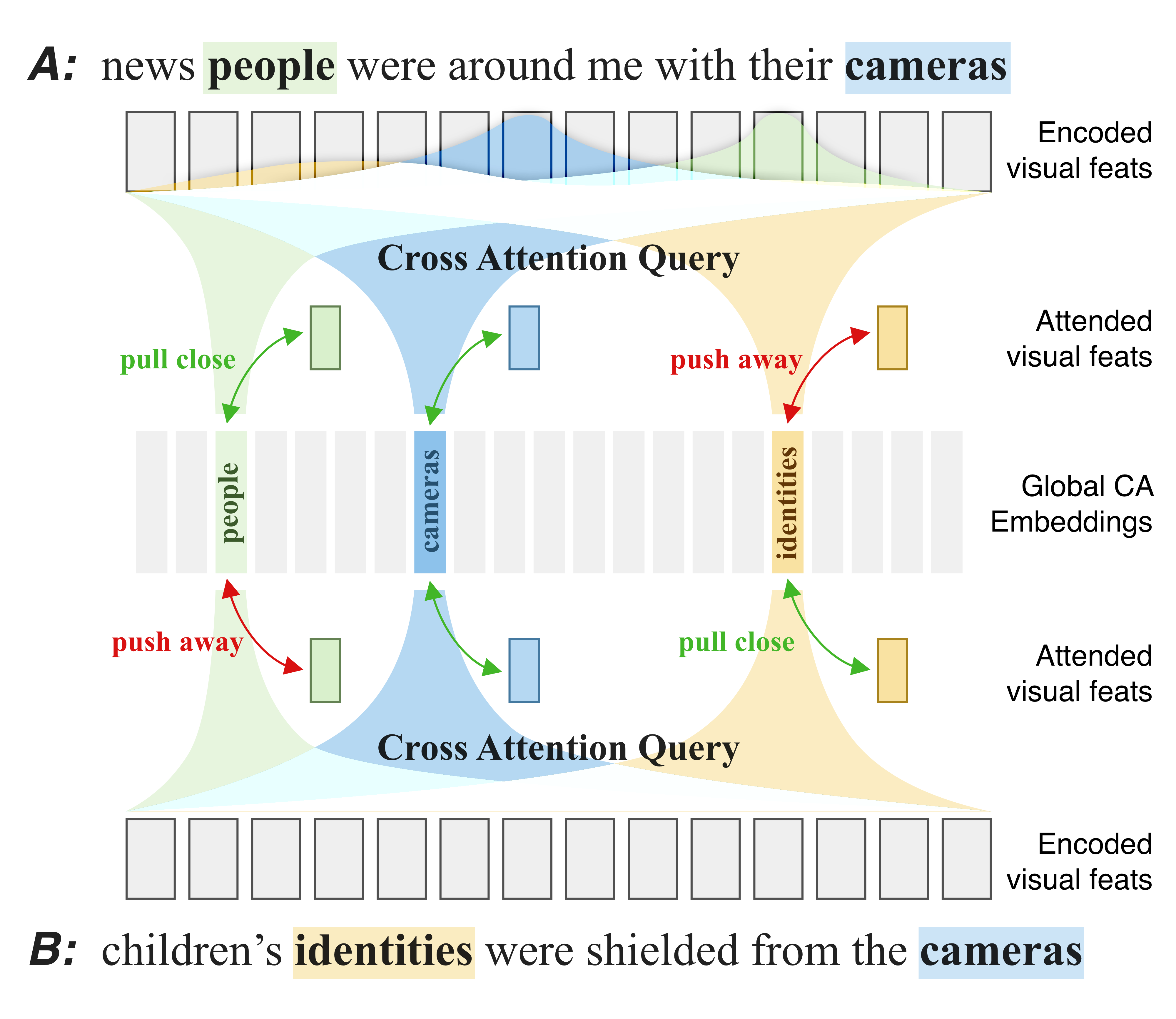}
    \caption{\label{fig:heading}Use global embeddings of conceptual words (CA, conceptual anchor) in spoken translation to supervise the visual feature instead of gloss. \textbf{A} and \textbf{B} are different samples from the same mini batch.}
    \vspace{-0.05in}
\end{figure}

In this paper, we study the task of sign language translation (SLT), which translates the sign videos into the corresponding spoken language. Glosses are the transliteration system for sign language. They serve as an intermediate representation of the signs. However, the vocabulary of gloss does not align with the spoken language nor does the order of the glosses. Unlike translation between two spoken languages, the number of frames in a sign video is much larger than the number of words in the spoken translation. This imposes a unique challenge for SLT. Models need to learn a clustering of the frames into gloss-level representation before they can translate the tokens. Previous methods solve this problem in two major ways, i.e., pre-train the visual backbone with gloss~\cite{camgoz2020sign} or jointly train on both translation and continuous recognition task~\cite{camgoz2020sign, chen2022simple} with an additional CTC loss~\cite{graves2006connectionist}. These methods have been proven effective, but the reliance on gloss annotations makes them hard to apply to more realistic scenarios. As gloss annotations require expert knowledge to make and often are limited in quantity or coverage of domains. Like the most frequently used PHOENIX14\textbf{T} dataset~\cite{camgoz2018neural} that focuses on weather reports or the KETI dataset~\cite{ko2019neural} that dedicates to emergencies. Datasets like OpenASL~\cite{shi2022open} and How2Sign~\cite{duarte2021how2sign} provide more samples but there are no gloss annotations for training. 

Motivated by these observations and the availability of large-scale SLT datasets, we designed a new framework that is gloss-free throughout the entire process and train the visual backbone jointly in an end-to-end manner.
The core idea of our method is illustrated in Figure~\ref{fig:heading}, we extract conceptual words from the ground truth spoken translation to be used as a weak form of intermediate representations. This exploits the shared semantics between signs and text. Though the extracted words might be different from the glosses, the concept expressed by these words should exist in both sign and text. We treat these words as conceptual anchors (CA) between the two modalities. Specifically, we use pre-trained GloVe embeddings~\cite{pennington2014glove} as the initialization of these anchors. Then they are treated as the query of cross attention against the encoded visual features.  As illustrated in Figure~\ref{fig:heading}, the query attend to each visual feature across the temporal dimension to calculate the similarity between the query and the visual features. With these similarities as weights of pooling, we get the attended visual features. The order of the most relevant features from the signing video does not match the order of the queries in the translation, so CTC is not viable in this situation. Instead, we impose the conceptual constraints in a contrastive manner. For each anchor word, we treated samples containing such words as positive and vice versus. For example, for the word \texttt{identities} in Figure~\ref{fig:heading} sample B is positive and sample A is negative. Query results for these positive and negative pairs along with the anchor word form a triplet, among which we conduct a hinge-based triplet loss. This process forces the visual2text encoder to learn the relation between different frames that is part of one sign. In all, our contribution can be summarized as:
\begin{itemize}
    \item An end-to-end sign language translation framework that takes the visual backbone in its training process. And we prove that proper design to accompany the text generation objective, will improve the performance of the framework rather than deteriorate it. 
    \item A replacement for gloss as a weak form of intermediate representation that facilitates the training of the visual backbone and encoder. It exploits the shared semantics between sign and text, bridging the gap between these two modalities. This also allows us to train the model on larger datasets without gloss annotations.
    \item We obtained state-of-the-art performance on the currently largest SLT dataset publicly available, improving the more modern BLEURT metric by a margin of $5.26$, which is $16.9\%$ higher than the previous state-of-the-art.
\end{itemize}
\section{Related Work}

\subsection{Sign Language Translation}
\textbf{Sign Language Translation:} Sign language translation (SLT) aims to translate a sign video containing multiple signs to the corresponding spoken text. \citet{camgoz2018neural} first proposed the PHOENIX14\textbf{T} dataset that enables the study of direct translation from sign videos to spoken translation. Due to the data scarcity issues caused by the cost of labeling gloss in large quantities. Most works on SLT focus on exploit gloss annotations~\cite{camgoz2020sign} or techniques like back translation~\cite{zhou2021improving} between gloss and spoken text. \citet{chen2022simple} transfers powerful pre-trained models~\cite{radford2019language, liu2020multilingual} to the sign domain through progressively pre-training and a mapper network. PET~\cite{jin-etal-2022-prior} utilizes the part-of-speech tag as prior knowledge to guide the text generation. However, they all rely on gloss annotations. There have been attempts to conduct SLT in a gloss-free manner~\cite{camgoz2018neural, li2020tspnet, kim2022keypoint}, but their results are subpar compared to those that use gloss annotation. Recently, there have emerged large-scale SLT datasets like How2Sign~\cite{duarte2021how2sign} and OpenASL~\cite{shi2022open}. They both surpass PHOENIX14\textbf{T} in quantity and are not limited to a certain domain. However, these two datasets don't provide gloss annotations. By far, there are few frameworks have been developed to tackle this challenging scenario except for the baseline methods of the datasets. 

\subsection{Pretraining with 
Weakly Paired Data}
Vision-language pretraining~\cite{radford2021learning,tan2019lxmert,chen2020uniter} on massive-scale weakly paired image-text data has recently achieved rapid progress. 
It has been proven that 
transferable cross-modal representations bring significant gains on downstream tasks~\cite{ri-tsuruoka-2022-pretraining,ling-etal-2022-vision,agrawal-etal-2022-vision}.
Recent endeavors~\cite{yu2022coca,desai2021virtex,wang2021simvlm,seo2022end} leverage generative pretraining tasks like captioning to enable the cross-modal generation capability.
Such a training regime has become increasingly popular in sign language translation.
In particular, a few early attempts~\cite{kim2022keypoint} directly adopted the translation loss for cross-modal learning. However, the translation objective is hard to learn an effective representation of the important concept, especially in an open domain scenario. In contrast, we design a contrastive concept mining scheme to address this problem, leading to performance gains on the two largest sign language translation datasets.


\section{Method}

\begin{figure*}[htp]
    \centering
    \includegraphics[width=\linewidth]{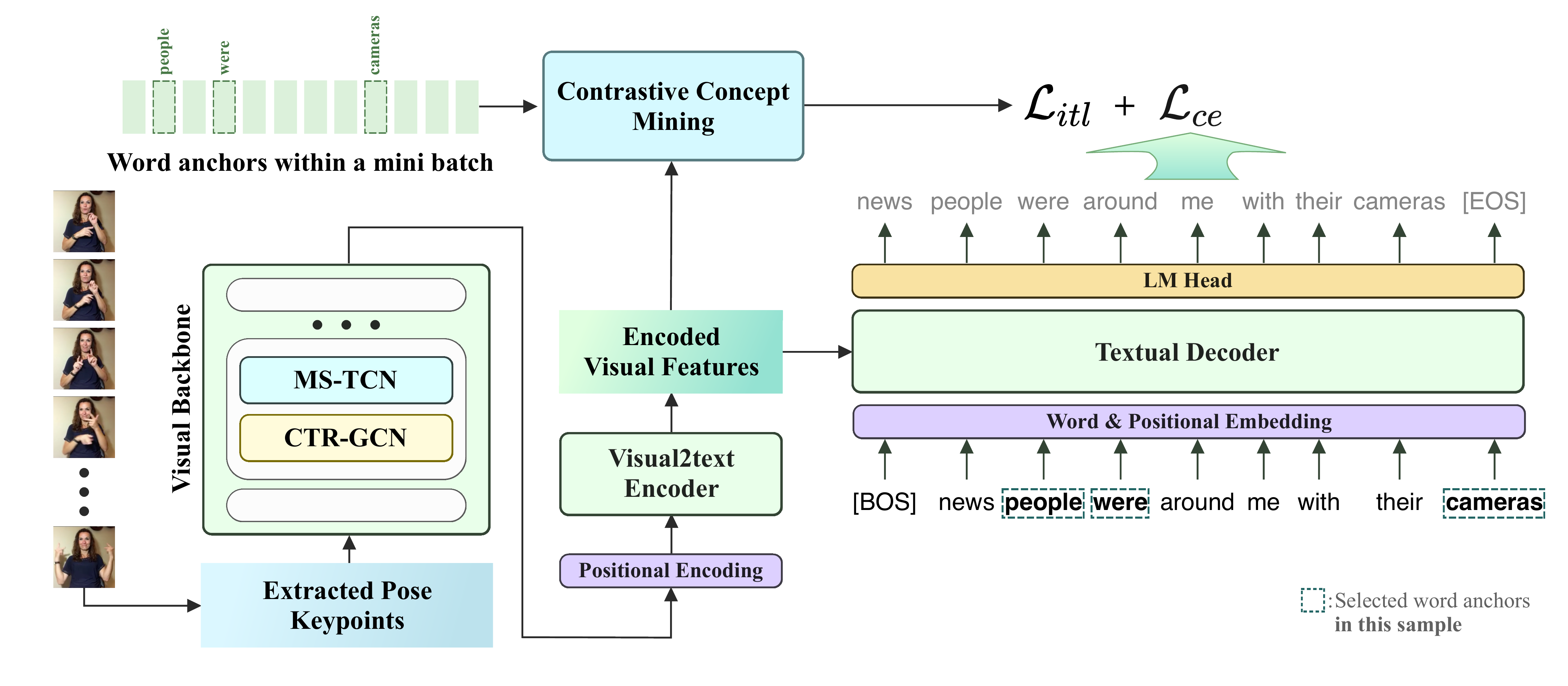}
    \caption{\label{fig:framework}Overall illustration of the framework.}
\end{figure*}

Given a sign video $X = \{f_1, f_2, \ldots, f_T\}$ of $T$ frames, our objective is to generate a spoken language sentence $Y = \{w_1, w_2, \ldots, w_L\}$ of length $L$ under the conditional probability $p(Y|X)$. Generally speaking, it holds that $T \gg L$. This trait makes the task of sign language translation harder compared to the translation task between different spoken languages. Past methods mostly use gloss supervision via CTC loss to impose an indirect clustering on the processed visual tokens. Gloss annotation provides the relative order and type of the signed word, not including the boundary between sign words. However, the making process of gloss annotations is labor-intensive, thus often in limited quantities. This restricts the scale of SLT datasets with gloss annotations. 

To this end, we are motivated to design a framework that can be trained only on sign video and translation pairs. To reduce processing load and translate longer sign vidoes, we extract pose landmarks $X_{pose} = \{p_1, p_2, \ldots, p_T\}$ offline from $X$ and use it as the input of our framework. In this section, we first give an overview of the proposed gloss-free end-to-end sign language translation framework, with details about each component. Then we elaborate on our approach aims to provide similar supervision to gloss in a self-supervised manner. 

\subsection{Framework Overview}

The overall structure of our framework is illustrated in Figure~\ref{fig:framework}. It consists of a modified CTR-GCN~\cite{chen2021channel} based visual backbone and a Transformer~\cite{vaswani2017attention} that takes in the visual features and generates the spoken translation.

\textbf{Frame Pre-processing}: To achieve end-to-end training on long video sequences, we choose to use pose keypoints extracted using MMPose~\cite{mmpose2020} as the input of our framework. This reduces the pressure on computing resources and enables us to use longer sequences of frames. Previous methods~\cite{li2020tspnet, camgoz2020sign} mostly rely on pre-processed visual features extracted using models like I3D~\cite{carreira2017quo} or CNN-based methods~\cite{szegedy2017inception, tan2019efficientnet}. It has also been proved in this work~\cite{camgoz2020sign} that a proper pre-training of the visual backbone can bring tremendous performance gain for the translation task. Then, it is a natural idea that we want to further improve the visual backbone through the supervision of the translation task. So we choose to use a lightweight GCN as our visual backbone and train the backbone all together.

\textbf{Visual Backbone}: The visual backbone takes in $T \times 76 \times 3$ keypoints including face, both hands and upper body. Each point contains $3$ channels, which indicates the 2D position and a confidence value ranging from $0$ to $1.0$. The output feature of all the keypoints is pooled by regions at the end of the network and produces a 1024-dimensional feature. The multi-scale TCNs~\cite{liu2020disentangling} in the backbone downsample the temporal dimension by a factor of 4. The backbone is pre-trained on the WLASL dataset~\cite{li2020word} through the isolated sign language recognition task.

\textbf{Visual2Text Encoder}: The visual2text encoder receives features from the visual backbone and translates these features from visual space to text space features $F_{enc} = \{s_1, s_2, \ldots, s_N\}$. It provides context for the textual decoder and the encoded visual features are also passed to the contrastive concept mining module. The output visual features of the visual backbone are combined with a fixed sinuous position encoding following~\cite{vaswani2017attention}, which provides temporal information for the encoder. 

\textbf{Textual Decoder}: The textual decoder models the spoken translation in an auto-regressive manner. During the training phase, the spoken translation target $Y$ is first tokenized using a BPE tokenizer~\cite{sennrich-etal-2016-neural} into $\hat{Y} = \hat{w}_{1:\hat{L}}$, which reduces out of vocabulary words during generation. Then we insert $\hat{w}_{0} = \texttt{[BOS]}$ and $\hat{w}_{\hat{L}+1} = \texttt{[EOS]}$ at the start and end to indicate the beginning and end of the decoding process. The tokens $\hat{Y}$ are converted into vectors through a word embedding layer and learned positional embedding, which is then summed together element-wise. Followed by layer normalization~\cite{ba2016layer} and dropout~\cite{srivastava2014dropout}. Then these vectors are passed through multiple transformer decoder layers to generate the feature  $F_{dec}= \{r_0, r_1, \ldots, r_{\hat{L} + 1}\}$ for each token. The vectors are masked to ensure causality, one token can only interact with tokens that came beforehand. We share the learned word embedding weights with the language modeling head at the end of the decoder similar to~\cite{press-wolf-2017-using, desai2021virtex}. 

\subsection{Cross-entropy Loss for Sign Translation}

The language modeling head $\mathcal{F}_{lm}$ in the textual decoder predicts probabilities over the token vocabulary. 
\begin{equation}
    p(x_{i}|x_{0:i-1}, F_{enc}) = \operatorname{softmax}(\mathcal{F}_{lm}(r_{0:i-1}))
\end{equation}
where $x_{i}$ indicates the hypnosis's $i_{th}$ token. Following previous literature on SLT, we use a cross-entropy loss at the training stage to supervise the text generation process. We have:
\begin{equation}
    \mathcal{L}_{ce} = - \sum_{i=0}^{\hat{L}}log(p(x_{i}|x_{0:i-1}, F_{enc}))
\end{equation}
This might be adequate for the translation of text pairs. Because for translating two text-based language inputs, the number of words for the text pair is similar (and there is no visual backbone too). But the number of frames of a sign video is much greater than either the number of corresponding glosses or spoken translation. It is very difficult for the encoder to learn a good representation as the token number of the encoder is much larger than that of the decoder, not to mention that we also want the encoder to provide good supervision for the visual backbone. In the work of~\citet{shi2022open}, they observed deteriorated performance if they tried to train the visual backbone and the transformer together. Thus we reckon in this case, a single cross-entropy loss at the end of the framework is not competent for our intended purpose. 


\subsection{Contrastive Concept Mining}

\begin{figure}[tp]
    \centering
    \includegraphics[width=\linewidth]{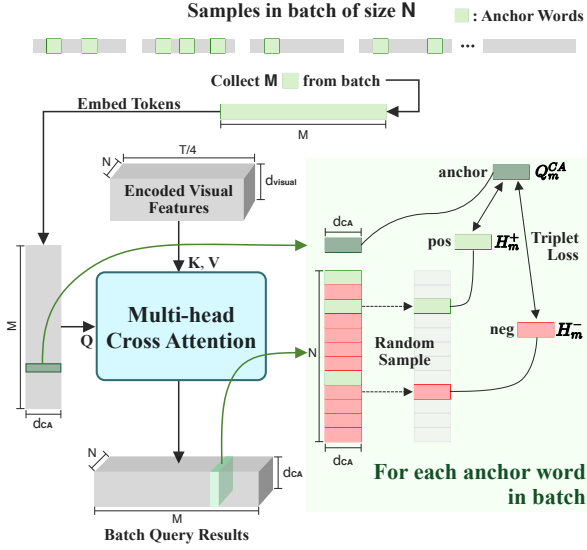}
    \caption{\label{fig:ccm}Overall illustration of the contrastive concept mining (CCM) process.}
\end{figure}

Under the presumption that single cross-entropy loss is not enough. We want to provide additional supervision for the visual2text encoder. We intend to achieve such effect by exploiting the shared semantics between sign and text. A sign video can be roughly considered as multiple chunks (ignoring transition between signs), with each chunk of consecutive frames representing one sign word (a gloss). Though we cannot get the exact sign word for each chunk as the spoken translation does not necessarily contains all the sign words and the orders also do not match. Key concepts expressed through sign and spoken translation should share the same underlying latent space. With this in mind, we design \textbf{C}ontrastive \textbf{C}oncept \textbf{M}ining (CCM) as shown in Figure~\ref{fig:ccm}.

The process of CCM consists of two steps: 1) Find possible words to be used as \textbf{C}onceptual \textbf{A}nchors(CA) in the training corpus, which we also refer to as anchor words. In practice, we mostly focus on verbs and nouns as we reckon such concepts are expressed in both the sign representation and the spoken language. It is natural to use these words as anchors for the encoder to structure the visual representations. 2) For each training batch of $N$ samples, we collect all the anchor words (total of $M$ words) in its spoken translations. For each word, we treat the sample containing such word as a positive sample and samples that do not as negative samples. Along with the global learned embedding for this word we conduct a triplet loss.

\textbf{Global CA query on encoded feats}: For a batch $B = \{x_1, x_2, \ldots, x_N\}$ of $N$ samples, we denote the collected word tokens as $B_{CA} = \{v_1, v_2, \ldots, v_{M}\}$. $M$ is the number of collected anchor words within the mini batch. These tokens are passed through an embedding layer to produce the query vector for multi-head cross attention. 
\begin{equation}
    Q^{CA} = EmbeddingCA(B_{CA})
\end{equation}
where $Q^{CA} \in \mathbb{R}^{M \times d_{ca}}$, $d_{ca}$ is the dimension of the embedding layer for conceptual anchors. For output features of the encoder $F_{enc}=\{s_1, s_2, \ldots, s_N\}$, in which $s_n \in \mathbb{R}^{L_{enc} \times d_{visual}}$. $L_{enc}$ represents the max token length output by the encoder, $d_{visual}$ is the dimension of the visual feature. The multi-head cross attention is defined as:
\begin{equation}
\begin{split}
    CrossAtten(Q^{CA}, s_n) = [head_1|\ldots|head_h]W^{o}& \\
    head_i = Attention(Q^{CA}W_{i}^{Q}, s_nW_{i}^{K}, s_nW_{i}^{V})&
\end{split}
\end{equation}
where $[.|.]$ denotes the concatenation operation, $head_i$ represents the output of the $i\text{-th}$ head. The projection matrices are $W_{i}^{Q} \in \mathbb{R}^{d_{ca} \times d}$, $W_{i}^{K}, W_{i}^{V} \in \mathbb{R}^{d_{visual} \times d}$ and $W^{o} \in \mathbb{R}^{hd \times d_{CA}}$, in which $d$ is the hidden dimension of the attention and $d_{CA}$ is the final output dimension(same as the embedding dimension for CA).The attention process is defined as:
\begin{equation}
    Attention(Q, K, V) = \operatorname{softmax}\left(\frac{Q K^\mathbf{T}}{\sqrt{d}}\right) V
\end{equation}
This process is repeated for each feature in $F_{enc}$. We denote $H_{n} = CrossAtten(Q^{CA}, s_n)$, we stack $\{H_{n}|n \in N\}$ to get the final output $H$ of cross attention. We have $H_{n} \in \mathbb{R}^{M \times d_{CA}}$ and $H \in \mathbb{R}^{M \times N \times d_{CA}}$.

The cross-attention operation finds the most relevant part of an encoded visual feature to the CA query. The embedding of these word anchors $Q^{CA}$ is shared across all the samples in the training set and is updated through the back-propagation process. We initialize these embeddings using pre-trained GloVe vectors~\cite{pennington2014glove}. The query results are the foundation for CCM, as we can encourage the encoder to gather visual information close to the word anchors and suppress noises similar to anchors but the anchor words are not presented in the sample.

\textbf{Inter-sample triplet loss}: We use a hinge-based triplet loss~\cite{wang2014learning} as the learning objective for the query results $H$. The selection of positive and negative samples is carried out within a mini batch. For each unique CA $v_m$ in a batch, we regard samples that contain this particular anchor word as positives and those that do not as negative samples. Since there might be more than one positive or negative sample for $v_m$, one positive or negative sample is chosen randomly. The objective function is formulated as:

\begin{equation}
\begin{split}
    l_{m} = \mu - &sim(H^{+}_{m}, Q^{CA}_{m}) + sim(H^{-}_{m}, Q^{CA}_{m}) \\
    &\mathcal{L}_{itl} = max(0, \frac{1}{M}\sum_{m=1}^{M}l_{m})
\end{split}    
\end{equation}

where $H^{+}_{m}$ and $ H^{-}_{m}$ denotes the query results for the sampled positive and negative sample for $v_m$ respectively. We use $sim(,)$ to calculate the cosine similarity as the distance between two vectors. $\mu$ is the margin for the triplet loss, it determines the gap between the distances of $H^{+}_{m}$ and $ H^{-}_{m}$ to the anchor $Q^{CA}_{m}$. 

\subsection{Training and Inference}
Our framework is trained by the joint loss $\mathcal{L}$ of cross-entropy loss $\mathcal{L}_{ce}$ and conceptual contrastive loss $\mathcal{L}_{itl}$, which is formulated as:
\begin{equation}
    \mathcal{L} = \mathcal{L}_{ce} + \lambda\mathcal{L}_{itl}
\end{equation}
where $\lambda$ is the hyper parameter that determines the scale of the inter-sample triplet loss. CCM only works during the training phase, and does not introduce additional parameters for inference. 

\section{Experiments}

In this section we provide details on the datasets and translation protocol we follow. Along with quantitative and qualitative results on different benchmarks. We also give a deep analysis about the design components of our method.

\subsection{Dataset and Protocols}

\textbf{OpenASL}: OpenASL~\cite{shi2022open} is a large-scale American sign language dataset collected from online video sites. It covers a variety of domains with over 200 signers. With $98,417$ translation pairs it's the largest publicly available ASL translation dataset to date. $966$ and $975$ pairs are selected as validation and test sets respectively.


\textbf{How2Sign}: How2Sign~\cite{duarte2021how2sign} is a large-scale American sign language dataset. It contains multi-modality data including video, speech, English transcript, keypoints, and depth. The signing videos are multi-view and performed by signers in front of a green screen. There are $31,128$ training, $1,741$ validation, and $2,322$ test clips.

\textbf{Gloss-free Sign2Text}:  \textit{Sign2Text} directly translates from continuous sign videos to the corresponding spoken languages as proposed by \citet{camgoz2018neural}. Unlike previous works, we ditch the need for gloss annotations throughout the entire framework including the pre-training phase.

\textbf{Evaluation Metrics}: To evaluate the translation quality, we report BLEU score~\cite{papineni2002bleu} and ROUGE-L F1-Score~\cite{lin-2004-rouge} following \citet{camgoz2018neural}. Same as OpenASL, we also report BLEURT score~\cite{sellam-etal-2020-bleurt}. BLEURT is based on BERT~\cite{devlin-etal-2019-bert} and trained on rating data, it can correlate better with human judgments than BLEU and ROUGE.

\subsection{Implementation Details}

In our experiment, we use PyTorch~\cite{paszke2019pytorch} to train the model on NVIDIA A100s. We rely on PyTorch's implementation of Transformers to build the framework. We use byte pair encoding tokenizer provided by Hugginface's Transformers~\cite{wolf-etal-2020-transformers} library. The tokenizers are all trained from scratch on the training split of the corresponding datasets.

\textbf{Network Details}: We use multi-head attention with $4$ heads in all transformer layers. The feed forward dimension in the transformer layers is set to $1024$, and we use $4$ layers both for encoders and decoders. For both OpenASL and How2Sign, we set the input frame cap to $512$. The word embedding layer is trained from scratch with a dimension of $768$.

\textbf{Training \& Testing}: The model is trained using the AdamW~\cite{loshchilov2017decoupled} optimizer. We use a linear learning rate scheduler with $1000$ warm-up steps. The learning rate is $3\times10^{-4}$ with $400$ epochs for both OpenASL and How2Sign. The models on OpenASL are trained across $4$ GPUs with a batch size of $48$ on each process for about 4 days. For How2Sign the model is trained across $8$ GPUs with a batch size of $40$ per process. In the text generation phase, we follow the common practice and use beam search with a beam size of $5$.

\textbf{Selection of anchor words}: We rely on NLTK's~\cite{bird2009natural} default POS (part-of-speech) tagger to select words used in CCM. First, the training corpus is tokenized using NLTK's punkt tokenizer. Then we pass the tokens to the POS tagger and filter out tags classified as general verbs or nouns (NN, NNP, NNS, VB, VBD, VBG, VBN, VBP, VBZ). Finally, we filter the verbs and nouns by their appearance frequency in the corpus. Words with occurrence not exceeding $10$ or close to the total sample count are discarded in this process. 

\begin{table*}[t]

\centering
\begin{adjustbox}{max width=\linewidth}
\begin{tabular}{lcccccc|cccccc}
\toprule
   \multirow{2}{*}{Methods} & \multicolumn{6}{c|}{DEV} &\multicolumn{6}{c}{TEST} \\ 
   \cmidrule{2-13} 
  & {\small ROUGE} & {\small BLEU-1} & {\small BLEU-2} & {\small BLEU-3} & {\small BLEU-4} & {\small \textbf{BLEURT}} & {\small ROUGE} & {\small BLEU-1} & {\small BLEU-2} & {\small BLEU-3} & {\small BLEU-4} & {\small \textbf{BLEURT}} \\
\midrule
Conv-GRU & 16.25 & 16.72 & 8.95 & 6.31 & 4.82 & 25.36 & 16.10 & 16.11 & 8.85 & 6.18 & 4.58 & 25.65 \\
I3D-transformer & 18.88	& 18.26 &	10.26 &	7.17 &	5.60 & 29.17 & 18.64 &	18.31 &	10.15 &	7.19 &	5.66 &	28.82  \\ 
 OpenASL & 20.43 & 20.10 & 11.81 &	8.43 &	6.57 & 31.22 &	21.02 &	20.92 &	12.08 &	8.59 &	6.72	& 31.09  \\ 
 \midrule \midrule 
 GloFE-N (ours)& \textbf{21.63} & \textbf{21.78} & \textbf{13.35} &	\textbf{9.61} &	\textbf{7.51} & \textbf{37.30} & 21.23 & 20.49 & 12.27 & 8.76 & 6.82	& \textbf{36.68}  \\
GloFE-VN (ours)& 21.37 & 21.06 & 12.34 &	8.68 & 6.68 & 36.75 & \textbf{21.75} & \textbf{21.56} & \textbf{12.74} & \textbf{9.05} & \textbf{7.06} & 36.35 \\ 
 \bottomrule
\end{tabular}
\end{adjustbox}
\caption{Results on the OpenASL dataset. \textbf{N} represents the model trained using only nouns as anchor words and \textbf{VN} means the model is trained using both verbs and nouns as anchor words. }
\label{tab:main-result} 
\end{table*}





\begin{table*}[ht!]
    \centering
    \begin{adjustbox}{max width=\linewidth}
    \begin{tabular}{lcccccc|cccccc}
        \toprule
         \multirow{2}{*}{Methods}  &\multicolumn{6}{c|}{DEV} & \multicolumn{6}{c}{TEST} \\
         \cmidrule{2-13} 
                                   & {\small ROUGE} & {\small BLEU-1} & {\small BLEU-2} & {\small BLEU-3} & {\small BLEU-4} & {\small \textbf{BLEURT}} & {\small ROUGE} & {\small BLEU-1} & {\small BLEU-2} & {\small BLEU-3} & {\small BLEU-4} & {\small \textbf{BLEURT}} \\
         \midrule 
         Alvarez$\dagger$   &   -   & 17.73 & 7.94 & 4.13 & 2.24 &   -   &   -   & 17.40 & 7.69 & 3.97 & 2.21 &   -   \\
         GloFE-VN (ours)                & 12.98 & 15.21 & 7.38 & 4.07 & \textbf{2.37} & 30.95 & 12.61 & 14.94 & 7.27 & 3.93 &  \textbf{2.24} & 31.65 \\
        \bottomrule
    \end{tabular}
    \end{adjustbox}
    \caption{Results on the How2Sign dataset. $\dagger$: \citet{alvarezsign} used CNN model~\cite{koller2019weakly} pre-trained with GLOSS annotations to extract the visual features. }
    \label{tab:comp_how2}
\end{table*}

\subsection{Comparison with state-of-the-art}
We test our framework on OpenASL against the multi-cue baseline proposed in the paper, as shown in Table~\ref{tab:main-result}. The baseline method incorporates multiple streams of global, mouth, and hands features and relies on external models to conduct sign spotting and fingerspelling sign search. Our framework, both GloFE-N (using only nouns as anchor words) and GloFE-VN (using both verbs and nouns as anchor words) surpasses all the previous methods on all metrics. The improvement on BLEURT stands out with a margin of $5.26$ for GloFE-VN on the TEST set, which is $16.9\%$ higher compared to the previous state-of-the-art. As for BLEURT on the DEV set, GloFE-N improves more than GloFE-VN with a gap of $6.08$ over the previous state-of-the-art.

We obtain the best TEST result of $7.06$ B4 with our VN model and the best DEV result of $7.51$ B4 with the N model. Though the N model obtains significantly higher scores on the DEV set, results on the TEST set are lower than the VN model. The vocabulary size on N is close to VN ($4,238$ to $5,523$), but as the N model only uses nouns the word type is less diverse. The lack of diversity makes the model less generalized, and more likely to fit the DEV set as it contains more similar samples to the training set. 

We also test the framework on How2Sign. The results are shown in Table~\ref{tab:comp_how2}. We surpass the previous method on BLEU-4 but fall behind on the BLEU metric measuring smaller n-grams. The VN vocabulary size for How2Sign is around $2,000$ which is close to the number of test clips in How2Sign. Combined with the higher B4, it shows that our framework is better at generating short phrases. But the coverage of concepts is limited by the vocabulary size of the anchor words. 

\begin{table}[ht!]
    \centering
    \begin{adjustbox}{max width=\linewidth}
    \begin{tabular}{ccc|cccccc}
        \toprule
         E2E        &     PE     &   CCM   & RG & B@1 & B@2 & B@3 & B@4 & {\small \textbf{BLEURT}}  \\
         \midrule 
                    &            &            & 18.94 & 18.25 & 10.38 & 7.37 & 5.81 & 34.35 \\
                    & \checkmark & \checkmark & 20.24 & 19.20 & 11.10 & 7.82 & 6.05 & 35.39 \\
         \checkmark & \checkmark &            & 20.92 & 20.37 & 11.62 & 8.09 & 6.23 & 35.65 \\
         \checkmark &            & \checkmark & 20.48 & 19.71 & 11.48 & 8.20 & 6.46 & 35.96 \\
         \checkmark & \checkmark & \checkmark & \textbf{21.75} & \textbf{21.56} & \textbf{12.74} & \textbf{9.05} & \textbf{7.06} & \textbf{36.35}\\
        \bottomrule
    \end{tabular}
    \end{adjustbox}
    \caption{Ablation on OpenASL demonstrates the effect of our different components. \textbf{E2E:} Whether to conduct end-to-end training to train the visual backbone together. \textbf{PE:} Fixed sinuous positional encoding added to the input of visual2text encoder.  \textbf{CCM:} Whether to use Contrastive Concept Mining on the encoded visual features during the training phase. B@N represents the BLEU-N score, this also applies to tables that came after this.}
    \label{tab:abl_comp}
\end{table}

\subsection{Ablation Study}

\subsubsection{Effect of Components} \label{subsubsec:effect_of_comp}

We examine the effectiveness of different design components as shown in Table~\ref{tab:abl_comp}. Namely, we ablate on the effect of the E2E (end-to-end training), PE (positional encoding for visual features), and CCM (contrastive concept mining), respectively. As a baseline, we first train a model without the three components. Without E2E, even we add PE and CCM both to the framework. The improvement over baseline is only at $0.24$ B4. If we add E2E back, this gap is widened significantly to $1.25$ B4. This proves that our design can improve the visual backbone's ability to recognize signs composed of multiple frames. With E2E, we also validate the effectiveness of PE and CCM, respectively. First, they both improve on the baseline line with a perceptible margin. When comparing PE to CCM, CCM is more performant, with an improvement of $0.65$ B4 against $0.42$ B4 over the baseline.


\subsubsection{Type of Anchor Words}

\begin{table}[ht!]
    \centering
    \begin{adjustbox}{max width=\linewidth}
    \begin{tabular}{cc|ccccccc}
        \toprule
         Word Type  &     Vocab.   & RG & B@1 & B@2 & B@3 & B@4 & {\small \textbf{BLEURT}}  \\
         \midrule 
         V   & 1693 & 20.96 & 20.70 & 11.89 & 8.42 & 6.51 & 36.50\\
         N   & 4238 & 21.23 & 20.49 & 12.27 & 8.76 & 6.82 & \textbf{36.68}  \\
         VN  & 5523 & \textbf{21.75} & \textbf{21.56} & \textbf{12.74} & \textbf{9.05} & \textbf{7.06} & 36.35 \\
         VNA & 6726 & 20.85 & 20.22 & 11.88 & 8.44 & 6.63 & 35.90 \\
        \bottomrule
    \end{tabular}
    \end{adjustbox}
    \caption{Ablation on selecting different types of words as the conceptual anchors. \textbf{V} and \textbf{N} stands for verb and noun respectively. \textbf{A} stands for adverbs and adjectives. The sum of the vocab size of \textbf{V} and \textbf{N} individually is greater than \textbf{VN} because the word type can vary depending on its relative position in the sentence. }
    \label{tab:abl_vocab}
\end{table}


We study the type of words selected in this experiment. From Table~\ref{tab:abl_vocab} we can see that with V, N, and VN, model performance increase as the size of the vocabulary increases. But when we added A (adverbs and adjectives) to the vocab, the performance deteriorates by $0.43$ B4. This is because the vocabulary jump from V to VN (or V to N), the number of conceptual word increases significantly. But with the addition of A, the extra words consists of major decorative purposes, they add to existing concepts (adverbs and adjectives modify verbs and nouns respectively). The number of conceptual word does not increase, but there are more anchors to attend to in the CCM process, which increases the learning difficulty.

\subsubsection{Inter-sample Triplet Loss Weight}

\begin{figure}[htp]
    \centering
    \includegraphics[width=\linewidth]{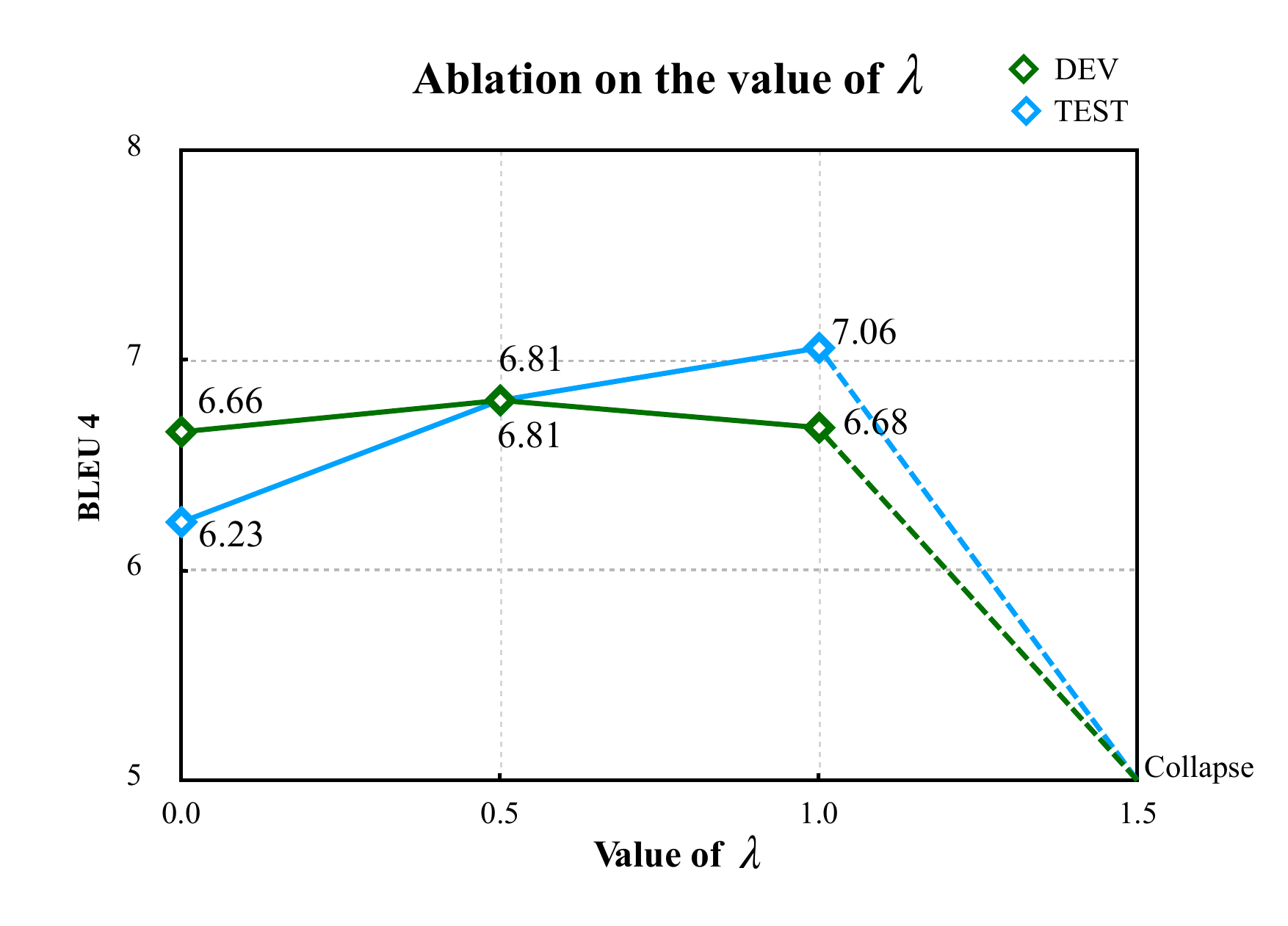}
    \caption{Ablation on the value of $\lambda$. Model collapsed when $\lambda$ is at $1.5$ or larger.}
    \label{fig:lambda}
\end{figure}
Here we study the effect of inter-sample triplet loss $\mathcal{L}_{itl}$ by varying the weight $\lambda$. As shown in Figure~\ref{fig:lambda}, B4 on the DEV set fluctuates within a small range while B4 on the TEST set increased $0.83$ as $\lambda$ increases from $0$ to $1.0$. The model collapsed after $\lambda$ goes beyond $1.5$. When $\lambda$ goes beyond $1.5$, $\mathcal{L}_{itl}$ takes the dominant spot in the combined loss. But $\mathcal{L}_{itl}$ alone cannot guide the generation process, resulting in the collapse of the model.



\subsection{Qualitative Results}









\begin{table}[ht!]
    \centering
    \begin{adjustbox}{max width=\linewidth}
    \begin{tabular}{|rp{1.0\linewidth}|}
        \hline
        Ref:        &    today is the first day of \textcolor{OliveGreen}{winter}.   \\
        Baseline:   &    today is the first day of the \textcolor{Red}{day}.   \\
        GloFE:      &    today is the first day of the \textcolor{OliveGreen}{winter} day.   \\
        \hline 
        \hline
        Ref:        &    \textcolor{OliveGreen}{meteorologists say} freeze warnings remain in the south including florida.   \\
        Baseline:   &    \textcolor{Red}{officials are warning} about 200 feet of snow.   \\
        GloFE:      &    \textcolor{OliveGreen}{meteorologists say} the weather will be keeping in louisiana.   \\
        \hline
        \hline
        Ref:        &    we have also reached out to ntid and asked for their \textcolor{OliveGreen}{response}.   \\
        Baseline:   &    we also reached out to the \textcolor{Red}{nad board members} for their \textcolor{Red}{stories}.   \\
        GloFE:      &    we also reached out to \textcolor{Red}{you} for their \textcolor{OliveGreen}{response}.   \\
        \hline
        \hline
        Ref:        &    the death toll from hurricane dorian is \textcolor{OliveGreen}{rising} in the \textcolor{OliveGreen}{bahamas}.   \\
        Baseline:   &    the death toll is \textcolor{Red}{now emotional}."   \\
        GloFE:      &    and the death toll in the \textcolor{OliveGreen}{bahamas} is \textcolor{OliveGreen}{rising}.   \\
        \hline
    \end{tabular}
    \end{adjustbox}
    \caption{Qualitative results of GloFE on the TEST split of OpenASL compared to the baseline model first introduced in~\ref{subsubsec:effect_of_comp}. We use \textcolor{Red}{red} to indicate mistranslation of conceptual words and \textcolor{OliveGreen}{green} to show the matching concepts between GloFE and Ref.}
    \label{tab:quali}
\end{table}

Table~\ref{tab:quali} shows the qualitative results of the generated text of GloFE compared to the baseline model. We mainly focus on whether the model generates the same conceptual words (verbs and nouns) as the reference text. For each sample, we show the reference text and the text generated by the baseline model and GloFE. We use red to indicate the mistranslated conceptual words in the baseline results and green to show the matching concepts. In the first example, both baseline and GloFE generate similar text with one key difference. GloFE successfully captures the concept of \texttt{winter} (noun) in the sign expression while the baseline does not. However, GloFE cannot always capture all the correct concepts. In the third example, GloFE failed the capture \texttt{ntid} and \texttt{asked}. But compared to the baseline, GloFE still managed to translate \texttt{response} correctly. In general, GloFE is capable of generating a more accurate translation of objects and motions expressed in the signing sequence.
\section{Conclusion}
In this paper, we propose a novel gloss-free end-to-end framework for sign language translation. Design an intermediate representation that can act as a fill-in when gloss annotation is not available. We exploit the shared semantics between sign and text, by extracting common conceptual words from the spoken translation. The model is trained end-to-end including the visual backbone, no gloss is used in training or pre-training, and achieves state-of-the-art performance on the largest sign languages translation dataset publicly available.

\section*{Limitations}

Our model is trained in an end-to-end manner, resulting in more training time costs than feature-based methods. 
To eliminate the need for gloss annotations, the CCM process relies on a large amount of sign and translation pairs. The generalizability of the model is restrained by the number of such pairs available.
The more ideal end-to-end framework should combine the visual backbone and visual2text encoder into one visual encoder that can be trained end-to-end. In addition, the selection of conceptual words is done according to manually-designed rules now and relies on external toolkits like NLTK. We will investigate automatic conceptual word extraction methods in future work. 

\section*{Ethics Statement}

Our work focuses on the task of sign language translation. Such systems aims to use technology to facilitate the day-to-day life of the deaf and hard-of-hearing community. Though we improve on the baseline, the proposed model still does not equip with the ability to serve as an interpreter in real-life scenarios. We use extracted keypoints as the input of the model, there are little to no concerns about personal privacy. For now, the model is only validated on American sign language datasets, currently it's not able to help people that do not use ASL.

\section*{Acknowledgements}
This work is supported by the Fundamental Research Funds for the Central Universities (No. 226-2022-00051).

\bibliography{anthology,custom}
\bibliographystyle{acl_natbib}

\end{document}